\documentclass{article}

\PassOptionsToPackage{numbers,sort&compress}{natbib}
\usepackage[main, final]{neurips_2026}

\usepackage[utf8]{inputenc} 
\usepackage[T1]{fontenc}    
\usepackage{hyperref}       
\usepackage{url}            
\usepackage{booktabs}       
\usepackage{amsfonts}       
\usepackage{nicefrac}       
\usepackage{microtype}      
\usepackage[table]{xcolor}
\usepackage{times}
\usepackage{soul}
\usepackage{url}
\usepackage[utf8]{inputenc}
\usepackage[small]{caption}
\usepackage{graphicx}
\usepackage{amsmath}
\usepackage{tabularx}
\usepackage{amsthm}
\usepackage{booktabs}
\usepackage{algorithm}
\usepackage{algpseudocode}
\usepackage{multirow}
\usepackage[table]{xcolor}
\usepackage{amsmath}
\usepackage{amssymb}
\usepackage{enumitem}
\usepackage{caption}
\usepackage{makecell}
\usepackage{tabularx}
\usepackage{ragged2e}
\definecolor{lightgreen}{rgb}{0.9, 1.0, 0.9}
\definecolor{lightblue}{RGB}{160, 180, 205}
\definecolor{APBEV}{HTML}{0F6CAD}
\definecolor{APTHREED}{HTML}{B54A3A}
\usepackage{graphicx}
\usepackage{subcaption}
\usepackage{amsthm}

\theoremstyle{definition}

\theoremstyle{plain}

\newcommand{\bestcell}[1]{\cellcolor{green!10}#1}
\usepackage{xspace}
\newcommand{\apbev}{\ensuremath{\mathrm{AP}_{\mathrm{BEV}}}\xspace}
\newcommand{\apthd}{\ensuremath{\mathrm{AP}_{\mathrm{3D}}}\xspace}

\title{Rethinking Radar 3D Perception: Representation Learning from Raw Spectra}

\author{%
Miao Zhang\textsuperscript{1,2}%
\thanks{These authors contributed equally.}  \quad
Mohamed Abdelsamad\textsuperscript{1,3}%
\footnotemark[1] \quad\
  Bin Yang\textsuperscript{1} \quad
  Sherif Abdulatif\textsuperscript{1} \\
  \textbf{Benedikt Loesch}\textsuperscript{1} \quad
  \textbf{Marco Altmann}\textsuperscript{1} \quad
  \textbf{Abhinav Valada}\textsuperscript{3} \quad
   \textbf{Bin Yang\textsuperscript{2}}\\[4pt]
\textsuperscript{1}Bosch Research\quad
\textsuperscript{2}University of Stuttgart \quad
\textsuperscript{3}University of Freiburg \\
\texttt{\{miao.zhang5,mohamed.abdelsamad,bin.yang3\}@de.bosch.com} \\
\texttt{valada@cs.uni-freiburg.de}\quad
\texttt{bin.yang@iss.uni-stuttgart.de} 
}

\begin{document}

\maketitle

\begin{abstract}

Raw radar spectra are widely viewed as too cluttered and noisy to support direct representation learning. This has led to handcrafted reductions, sparse point clouds, or compact Doppler descriptors as the default preprocessing step in radar perception. In this work, we challenge this assumption and argue that the main limitation is not the raw spectral representation itself, but rather the absence of a learning objective capable of separating the structured signal from noise. Handcrafted reductions partially address this problem, but they impose rigid assumptions about what counts as signal, discarding weak but informative structure while retaining artifacts that match the reduction model. Once such information is removed, supervised learning cannot recover it or question what was preserved. We propose a principled alternative: structure is the component of the radar spectrum that remains reconstructable under localized corruption, while noise is the component that does not. This perspective makes self-supervised masked modeling a natural learning paradigm for dense radar spectra, without imposing a predefined signal–noise separation. We instantiate it with Leakage-aware Spectral MAE, a masked autoencoding framework for raw radar cubes that incorporates a corruption strategy inspired by the cross-bin energy spreading caused by spectral leakage. We also show that Doppler organization is crucial: treating Doppler as a primary spatial dimension, using an x-y-D representation with height encoded as features, substantially outperforms naive channel stacking and is essential for exploiting dense 4D spectra. Without any handcrafted reductions, our approach consistently outperforms state-of-the-art radar-based detectors and point-cloud-based self-supervised learning (SSL) counterparts. These results establish raw radar spectra as a highly effective representation for 3D perception when paired with an appropriate pretraining objective.
\end{abstract}

\section{Introduction}

Radar sensors are becoming increasingly important to autonomous driving as they remain reliable  under adverse weather and lighting conditions, where cameras and LiDARs often degrade~\cite{schramm2024bevcar}. However, the rich frequency-domain spectra produced by radar hardware are rarely used directly in perception pipelines. Instead, they are first compressed through handcrafted reductions such as CFAR detection~\cite{rohling2007radar}, which collapses them into sparse point clouds, or compact statistics that summarize their Doppler content~\cite{palffy2022vod, 
liu2023smurf}. Modern radar perception systems are then developed around these reduced representations~\cite{ding2024radarocc, leitgeb2026rade, parit2026real, nayak2024ralf}, and substantial effort has gone into refining 
this paradigm~\cite{cheng2024centerradarnet, Kong2025RTNHplus, peng2024mufasa}. 

This pipeline has an important but largely unexamined cost. Handcrafted reductions do not merely suppress noise but they \textit{define} what will be treated as noise before learning begins. Weak reflections 
that fall below CFAR thresholds are discarded. Doppler structure not captured by summary 
statistics is discarded. The resulting representation is therefore not simply a cleaner version of the radar measurement. It is the subset of the measurement that survives a fixed set of assumptions about what signal should look like. Any weak, distributed, or noncanonical structure that violates these assumptions is removed, even if it is informative for downstream perception.

A natural alternative is to learn directly from unreduced post-FFT radar spectra, which we refer to as raw spectra because they preserve the spectral information otherwise discarded by handcrafted reductions. However, prior attempts have not established dense spectra as a clearly superior representation. Supervised learning on fully dense spectra under realistic driving conditions does not reliably outperform reduced inputs~\cite{paek2023enhanced}, a finding we also reproduce. This outcome is often interpreted as evidence that raw spectra are too noisy to support direct learning. We argue that the failure lies in the objective, not the input. Supervised perception losses, such as those based on bounding boxes or semantic labels, are defined over the task's output space. They provide only indirect supervision
for how the model should interpret the input. When trained on dense spectra, a supervised model is exposed to all components that handcrafted pipelines would have removed, but receives no explicit signal about which spectral structures are coherent and which are incidental.

This suggests that dense radar perception requires an objective that reasons about structure in the input space itself, without presupposing the morphology of signal or noise. We propose a novel definition: \textit{signal is what remains reconstructable under localized corruption, and noise is what does not}. An objective built on this definition requires no prior on signal or noise morphology. It discovers signal boundaries from the data itself. This is related to the $\mathcal{J}$-invariant framework of~\cite{batson2019noise2self}, but the problems are fundamentally different. Those methods assume an underlying clean signal and use
self-supervision to recover it from noisy observations. Raw radar spectra have no clean counterpart. Our goal is not to recover a hidden denoised spectrum, but to learn which structures in the observed spectrum are coherent enough to support prediction. This view makes self-supervised masked modeling a natural learning paradigm for dense radar spectra.

We instantiate this principle by introducing \textbf{Spectral MAE (SPMAE)}, the first masked autoencoding 
framework for raw radar spectra, and \textbf{Leakage-aware Spectral MAE (L-SPMAE)}, a physically grounded extension that corrupts the visible context by simulating the cross-bin energy spreading induced by missing pre-FFT measurements. We find that leakage corruption alone achieves performance comparable to random masking, and that the two yield the strongest representations. Our results also reveal that the organization of Doppler is critical. Simply stacking Doppler bins as channels brings no benefit over mean Doppler aggregation, while treating Doppler as a primary spatial dimension in an x-y-D representation substantially improves downstream perception after self-supervised pretraining.

\begin{figure}
    \centering
    \includegraphics[width=0.95\linewidth]{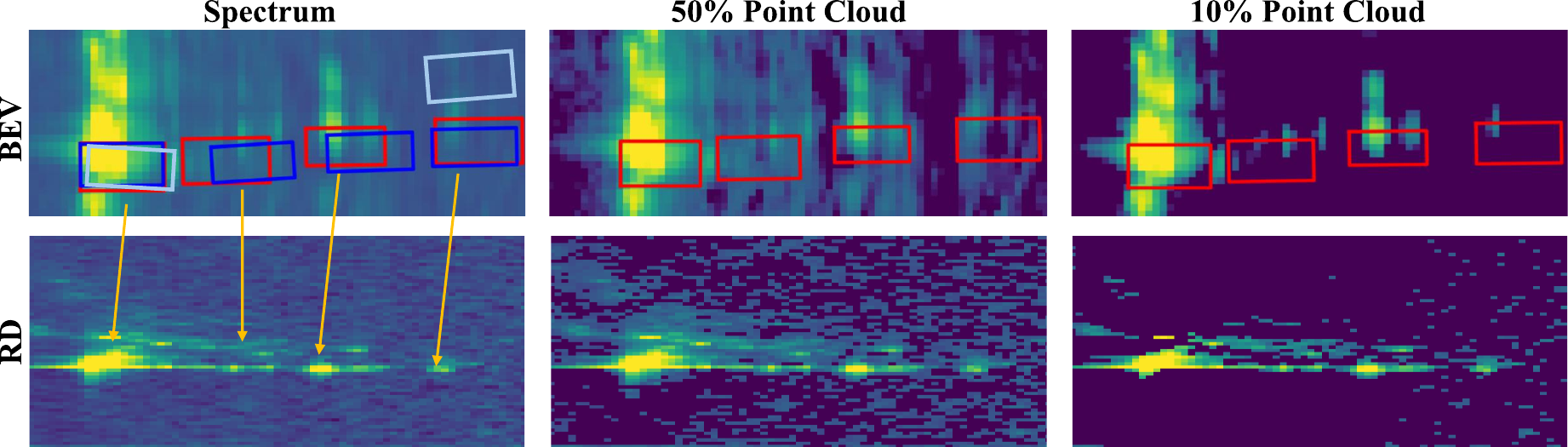}
\caption{Reducing dense spectra into sparse point representations suppresses both structured signals and noise simultaneously, which may remove informative spectral structures and degrade perception performance. In contrast, dense spectra preserve the full spectral information, whereas SSL enables the model to better exploit useful structures in noisy observations. \textcolor{red}{Red}: ground truth, \textcolor{blue}{blue}: prediction by spectrum-based SSL, \textcolor{lightblue}{light blue}: prediction by point-cloud-based SSL. RD: Range-Doppler view. BEV: Bird's-eye view.}
    \label{fig: leading}
    \vspace{-0.6cm}
\end{figure}

Figure~\ref{fig: leading} illustrates the central empirical finding: supervised learning degrades as spectral density increases, whereas self-supervised pretraining benefits from the additional spectral information. 
On K-Radar~\cite{kradar}, our approach surpasses recent radar-specific detectors
such as RTNH+~\cite{Kong2025RTNHplus} and RADENet~\cite{leitgeb2026rade}, despite
using a generic 3D detector~\cite{ptv3} rather than a radar-specific architecture.
Consistent improvements are also observed on 
RADIal~\cite{Radial}. These performance improvements are not explained by model capacity or additional supervision. They come from giving the model access to signals that handcrafted pipelines discard, together with an objective capable of discovering coherent structure within dense spectra.

Our contributions are as follows:
\begin{itemize}[noitemsep,topsep=0pt]
    \item We introduce a principle for learning from unprocessed radar spectra: signal is the component that remains reconstructable under localized corruption. This view motivates self-supervised masked modeling as a natural learning paradigm for dense radar spectra.

    
    \item Based on this insight, we introduce \textbf{SPMAE} for radar spectrum pretraining and further propose \textbf{L-SPMAE}, a physically grounded extension that simulates cross-bin spectral leakage induced by missing pre-FFT measurements, yielding stronger representations than random masking alone.

    \item An $(x,y,D)$ Doppler encoding that treats Doppler as a primary spatial dimension rather than a channel feature, which we show is necessary for SSL pretraining to exploit the full 4D radar cube.

    \item State-of-the-art 3D detection on K-Radar and RADIal using a generic 
    detector pretrained on raw spectra, surpassing radar-specific detectors and 
    SSL baselines trained on reduced representations, without any handcrafted 
    reduction.
\end{itemize}

\section{Related Work}
\label{sec: related}

\textbf{Radar Signal Representation}. 
\label{sec: rsp}
Radar systems acquire time-domain analog-to-digital converter (ADC) signals and apply DFT-based processing to obtain a 4D spectrum spanning range, azimuth, elevation, and Doppler~\cite{han20234d}, which is then followed by a signal processing reduction pipeline to obtain a sparse, clean representation, such as a point cloud. CFAR algorithms~\cite{rohling2007radar} serve as a standard approach for detecting peak reflections above an adaptive threshold. In addition, sparse radar point clouds in K-Radar~\cite{kradar} are generated by selecting the top $N\%$ reflections ranked by power. Furthermore, RTNH+~\cite{Kong2025RTNHplus} introduces a combined CFAR-based two-level preprocessing method to capture different radar characteristics. Recently, learnable radar representations have been explored~\cite{wang2025sddiff,roldan2024see}, but these approaches typically rely on supervision from other modalities such as LiDAR or camera.

\textbf{Radar 3D Perception}. 
Building upon these representations, radar sensors have been increasingly adopted for 3D perception tasks~\cite{zeller2023rit,palffy2022vod,cheng2024centerradarnet}, especially 3D object detection. Most existing methods extract statistical features from CFAR-filtered radar point clouds. For example, SMURF~\cite{liu2023smurf} uses kernel density estimation, while MUFASA~\cite{peng2024mufasa} designs local geometric descriptors based on Lalonde features. A few methods also operate directly on radar spectra, but still rely on compressed representations. For instance, RADENet~\cite{leitgeb2026rade} projects the 4D spectrum into 2D by compressing the Doppler and elevation channels. Overall, existing radar perception methods rely heavily on handcrafted reductions and sparse supervision, limiting their ability to exploit dense radar spectra. We argue that self-supervised learning directly on radar spectra offers a more suitable alternative.

\textbf{3D Self-Supervised Learning}. 
Reconstruction-based SSL has proven effective for learning geometric representations 
in 3D perception. GeoMAE~\cite{tian2023geomae} learns structure-aware features by 
reconstructing geometric properties such as centroids, OccupancyMAE~\cite{min2023occupancy} 
proposes voxel occupancy as a pretraining target, and NOMAE~\cite{nomae} extends this 
by predicting multi-scale occupancy in neighborhoods of visible tokens. Self-distillation 
approaches have also shown promising results~\cite{wu2025sonata,abdelsamad2026dos,yang2026pointins,abdelsamad2026ghostpoint}, 
although they typically rely on carefully designed augmentations or regularizations to avoid collapse, whereas 
reconstruction-based methods directly exploit the intrinsic structure of the input.
In radar perception, existing SSL works either operate on point clouds~\cite{ding2022raflow,
li2025super4dr}, depend on cross-modal supervision~\cite{hao2024bootstrapping,luo2025radler},
or apply MAE-style pretraining to RF signals with sparsity-aware masking, as in PRISM~\cite{fang2024prism}.
However, SSL tailored to the intrinsic structure of unreduced radar spectra remains largely underexplored. Our work is further 
inspired by Noise2Self~\cite{batson2019noise2self}, which shows that $\mathcal{J}$-invariant 
masking enables self-supervised prediction without clean supervision, a perspective 
particularly relevant to radar, where observations are inherently noisy.

\section{Methods}
In this section, we develop a framework for learning directly from raw radar spectra 
without any handcrafted reduction. Section~\ref{sec: theory} establishes the 
theoretical basis for signal discovery in dense \mbox{spectra} using corruption and reconstruction. 
We instantiate this through two complementary corruption strategies: 
Spectral MAE (Section~\ref{sec: spmae}), which applies standard masked autoencoding 
to raw radar spectra, and Leakage-aware Spectral MAE (Section~\ref{sec: splmae}), 
which extends it with a physically grounded corruption strategy. Both are developed 
under the RAE-cube formulation, with Doppler encoded as features. 
Section~\ref{sec: doppler_encoding} later shows that how Doppler enters this 
representation is itself a critical design choice, and introduces a RAD-cube 
reformulation that the framework accommodates without modification.

\subsection{Signal Discovery via Corrupted-Context Prediction}
\label{sec: theory}

The radar spectrum is a 4D tensor, but for the purpose of this treatment, we flatten the radar spectrum to a vector $X \in \mathbb{R}^m$. The partition 
structure we define is over indices and does not depend on any particular ordering, 
so the treatment applies to any indexing of the spectrum. Let $J \subset \{1, 
\ldots, m\}$ be a subset of indices and $J^c = \{1, \ldots, m\} \setminus J$ its 
complement. We define the family of $\mathcal{J}$-invariant functions as all functions $f$ 
whose output on $J$ does not depend on $X_J$. A network that receives only 
$X_{J^c}$ as input is by definition $\mathcal{J}$-invariant.

For a fixed partition $\{J, J^c\}$, the self-supervised reconstruction loss over 
the $\mathcal{J}$-invariant family is
\begin{equation}
    \mathcal{L}_J(f) = \mathbb{E}_X\left[\|f(X_{J^c})_J - X_J\|^2\right],
\end{equation}
whose unique minimizer is $f^*_J(X_{J^c})_J = \mathbb{E}[X_J \mid X_{J^c}]$. 
When $J$ and $J^c$ are statistically coherent, $X_{J^c}$ is informative about 
$X_J$ and $\mathcal{L}_J$ will be low at the optimum. When they are independent, 
the optimum reduces to the marginal mean $\mathbb{E}[X_J]$ and $\mathcal{L}_J$ 
will be high. The per-partition loss is therefore a measure of statistical 
coherence between $J$ and $J^c$.

The global loss takes an expectation over both $X$ and randomly drawn partitions $J$:
\begin{equation}
    \mathcal{L}(f) = \mathbb{E}_{X,J}\left[\|f(X_{J^c})_J - X_J\|^2\right].
\end{equation}
Since $J$ is sampled randomly, the model cannot rely on any fixed set of coherent 
indices always being visible. Across many training steps, coherences at all strengths 
and scales are alternately available in context and masked as targets, forcing the 
model to discover and represent the full coherence structure of the spectrum.

We define \textit{signal} as the coherence structure of $X$, operationalized as 
$f^*(X)$: the output of the model trained under the global loss when presented with 
the full unmasked spectrum at inference time. Indices statistically independent of 
all others do not reduce $\mathcal{L}_J$ under any partition and are not represented 
in $f^*$. Indices that participate in coherent structure are represented in 
proportion to that coherence.

Radar physics provides intuition for this notion of coherence. Target extent, the sensor’s point spread function, and spectral leakage spread reflected energy across neighboring bins, making target responses locally predictable, whereas thermal noise is less predictable from context. This predictability can also extend to multipath and stable clutter, which may carry useful information about objects and scene geometry. Learning these coherent components preserves that information for downstream fine-tuning to exploit. We stress that this interpretation is a motivating argument rather than a formal proof. We validate it empirically in Section 5.

This framework is related to the $\mathcal{J}$-invariant framework of 
Noise2Self~\cite{batson2019noise2self}, from which we borrow the loss formulation. 
The problems are, however, fundamentally different: Noise2Self assumes a clean signal 
exists and uses $\mathcal{J}$-invariance to recover it under a noise independence assumption, 
whereas we make no such assumption and define signal as the coherence structure of 
the spectrum itself. The corruption strategy need not be limited to binary masking: 
any transformation that renders $X_J$ uninformative about itself defines the same 
objective, and physically motivated corruptions that distort rather than remove can 
define harder and more structured prediction tasks.

\subsection{Spectral MAE}
\label{sec: spmae}

Following the corrupted-context formulation in Section~\ref{sec: theory}, we instantiate 
a self-supervised learning method termed Spectral MAE (SPMAE), which applies masked 
autoencoding directly to raw radar spectra.
Given a 4D radar spectrum $\mathbb{R}^{R \times A \times E \times D}$ indexed by range, azimuth, elevation, and  Doppler,
we convert the center of each $(r,a,e)$ bin from range--azimuth--elevation coordinates 
to Cartesian coordinates using the corresponding bin resolutions. This yields 
$P = R \times A \times E$ spectrum points with
\begin{equation}
 \mathbf{s}_i = (\mathbf{x}_i, \mathbf{f}_i),\mathbf{x}_i = (x_i, y_i, z_i), \quad S = \{\mathbf{s}_i\}_{i=1}^{P}.
\end{equation}
The feature $\mathbf{f}_i$ is obtained by aggregating along the Doppler dimension. In 
the absence of ego-motion information, we follow K-Radar~\cite{kradar} and use the mean 
Doppler power as the point feature, which we refer to as RAE.

During pretraining, we randomly mask a subset of spectrum points $S$ according to a 
mask ratio $m_{\mathrm{rate}}$. Let $S_v$ and $S_m$ denote the visible and masked 
spectrum points, respectively. The visible points $S_v$ are fed into the encoder $\mathcal{E}$ 
to obtain latent representations. For the masked positions indexed by $S_m$, we 
introduce a set of mask tokens $\{t_i\}_{i\in S_m}$ and pass them to the decoder $\mathcal{D}$
together with the encoded visible tokens. The training objective is to recover the 
original feature vectors $\mathbf{f}_i$ of the masked spectrum points:
\begin{equation}
\mathcal{L}_{\mathrm{SPMAE}}
= \frac{1}{|S_m|}\sum_{i \in S_m}
\left\|
\mathcal{D}\!\left(\mathcal{E}(S_v), t_i\right) - \mathbf{f}_i
\right\|_{1},
\label{eq: spmae_loss}
\end{equation}
where we use the $\ell_1$ loss and average it over masked locations.

\subsection{Leakage-aware Spectral MAE}
\label{sec: splmae}
\begin{figure}[t]
    \centering
    \includegraphics[width=\linewidth]{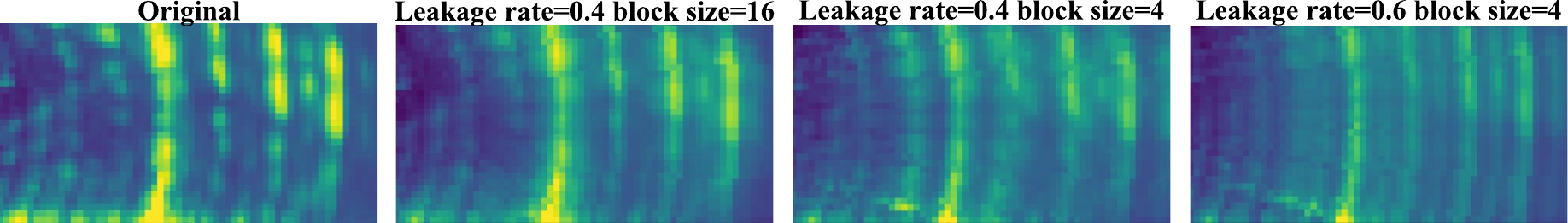}
\caption{BEV visualization of the leakage effect. A higher leakage rate $l_{\mathrm{rate}}$ leads to stronger energy spreading. Larger block sizes $b$ produce more structured and spatially correlated low-frequency distortions, while smaller blocks generate less correlated and more diffuse high-frequency perturbations.}
    \label{fig: leakage}
\end{figure}
\begin{figure*}[t]
    \centering
    \includegraphics[width=0.95\linewidth]{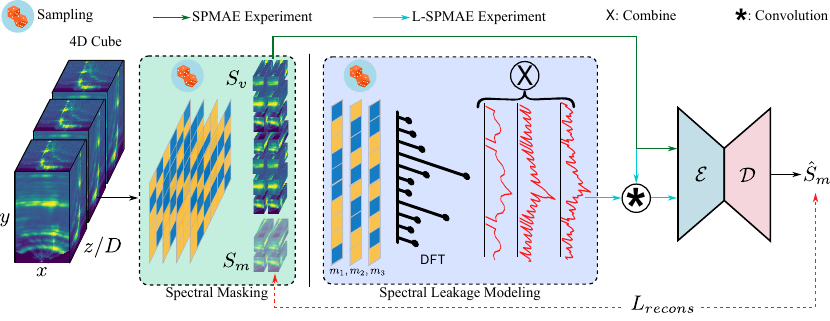}
\caption{Overall pipeline. For \textbf{L-SPMAE}, random spatial masking is applied to the radar spectrum $S$, producing the visible spectrum $S_v$ and masked spectrum $S_m$. Spectral leakage modeling is then applied to $S_v$ (Section~\ref{sec: splmae}) to generate the corrupted $\tilde{S}_v$, which is fed into the encoder $\mathcal{E}$, followed by the decoder $\mathcal{D}$ with mask tokens for reconstruction. For downstream tasks, only $\mathcal{E}$ is retained as the feature extractor.}
    \label{fig: Methods}
\vspace{-5pt}
\end{figure*}

Although SPMAE provides a simple masked reconstruction baseline, random spatial masking ignores a key property of radar spectra: spectral bins are coupled through the DFT. In digital signal processing, finite-length sampling is equivalent to applying a window in the signal domain, whose frequency response induces spectral leakage~\cite{lyon2009discrete}. For a signal $z$ and a binary sampling window $m$, this can be written as
\begin{equation}
\mathrm{DFT}(z \odot m)
=
\mathrm{DFT}(z) \circledast \mathrm{DFT}(m),
\label{eq: leakage}
\end{equation}
where $\odot$ denotes element-wise multiplication and $\circledast$ denotes circular convolution. Thus, windowed measurements do not merely remove isolated spectral bins but also cause cross-bin energy spreading in the spectral domain.

Motivated by this observation, we propose Leakage-aware Spectral MAE (L-SPMAE), a radar-specific extension of SPMAE that directly models leakage effects on radar spectra. Rather than corrupting spectrum points independently, we generate random binary windows and transform their DFT responses into controllable convolution kernels that simulate cross-bin energy spreading in the spectral domain. The resulting corrupted views better capture radar-specific spectral distortions and provide a physically inspired pretext task distinct from standard point masking. The randomized corruption is designed to make reconstruction challenging, rather than to match the statistics of actual hardware leakage.

To construct a leakage-aware corrupted 
view, we generate axis-wise binary windows
\begin{equation}
\begin{aligned}
m_1 &\in \{0,1\}^{R},~
m_2 \in \{0,1\}^{A},~
m_3 \in \{0,1\}^{E},
\end{aligned}
\end{equation}
where each window keeps a fraction $(1-l_{\mathrm{rate}})$ of samples along the 
corresponding axis. The leakage rate $l_{\mathrm{rate}}\in[0,1]$ controls the strength 
of the induced corruption: a larger $l_{\mathrm{rate}}$ preserves fewer effective 
samples and therefore produces stronger energy spreading. To generate more structured leakage patterns, we sample the binary windows in contiguous 
blocks with a block size $b$. Larger blocks introduce more structured and spatially 
correlated perturbations, resulting in more localized and low-frequency distortions in 
the spectrum domain. In contrast, smaller blocks produce less correlated, high-frequency 
perturbations, which lead to more diffuse and globally distributed cross-bin energy 
spreading.  We then compute a separable 
leakage kernel
\begin{equation}
K =
\left|\mathrm{DFT}(m_1)\right|^2
\otimes
\left|\mathrm{DFT}(m_2)\right|^2
\otimes
\left|\mathrm{DFT}(m_3)\right|^2,
\end{equation}
where $\otimes$ denotes the outer product. The squared magnitude is used to model energy 
spreading in the spectrum domain.  The corrupted spectrum is obtained by circular 
convolution $\circledast$ along the three spatial axes independently for each channel. An example of the corrupted spectrum is visualized in Figure~\ref{fig: leakage}.

As shown in Figure~\ref{fig: Methods}, L-SPMAE combines leakage-aware corruption with masked reconstruction in a decoupled manner. Random masking is first applied to $S$ to obtain the visible spectrum $S_v$, after which the leakage kernel $K$ is used to construct the corrupted spectrum:
\begin{equation}
\tilde{S}_v
=
S_v \circledast K.
\label{eq: leakage_corruption}
\end{equation}

The reconstruction target remains the original $\mathbf{f}_i$ from the uncorrupted masked spectrum $S_m$. The only difference from Equation~\ref{eq: spmae_loss} is that $\tilde{S}_v$ is used as input instead of the original $S_v$. This design makes the reconstruction task more radar-consistent than SPMAE, since the model must infer missing targets not only from incomplete observations, but also from contexts distorted by cross-bin leakage. Further details of the leakage modeling are provided in Appendix~\ref{appendix: algo}.


\subsection{Doppler-aware Spectrum Encoding}
\label{sec: doppler_encoding}

The framework developed in Sections~\ref{sec: spmae} and~\ref{sec: splmae} uses the 
RAE-cube formulation, in which Doppler is aggregated into point features via mean 
Doppler power. A natural question is whether SSL can extract additional value from 
the full 4D spectrum by retaining all Doppler bins. The obvious approach is channel 
stacking (which we refer to as RAE-d), but we find it brings no benefit over mean Doppler aggregation: without a 
spatial axis for Doppler, the encoder cannot model Doppler structure, and SSL 
pretraining cannot leverage Doppler coherence as a predictive signal.

To address this, we introduce a Doppler-aware reformulation that treats Doppler as a 
primary spatial dimension. We swap the elevation and Doppler axes and construct 
spectrum points from $(r,a,d)$ bins, yielding $P = R \times A \times D$ points with
\begin{equation}
\mathbf{x}_i = (x_i, y_i, d_i),
\end{equation}
where $(x_i, y_i)$ are obtained from range and azimuth bins and $d_i$ denotes the 
Doppler coordinate (denoted as RAD). The associated feature vector is taken from the elevation dimension:
\begin{equation}
\mathbf{f}_i \in \mathbb{R}^{E}.
\end{equation}
Height is encoded as channels while Doppler becomes an explicit axis for attention, 
allowing the encoder to directly model dependencies along the Doppler dimension. 
SPMAE and L-SPMAE apply without modification: $\mathbf{x}_i$ changes from $(x,y,z)$ 
to $(x,y,d)$, $\mathbf{f}_i$ changes from mean Doppler power to elevation bins, and 
for L-SPMAE the dense grid becomes $\mathbb{R}^{R \times A \times D \times C}$ 
with $C = E$.

\section{Experiments}
\subsection{Dataset}
\label{sec: dataset}
\textbf{K-Radar~\cite{kradar}} is the largest open-source 4D radar dataset providing raw 4D radar spectrum data.
The dataset contains over 35k frames with object detection annotations.
As noted in previous works~\cite{fent2024dpft,zhang2024exploring,Zhang2025capmix}, the official split divides individual sequences into subsequences that appear in both the training and testing sets. Therefore,
in addition to the official split, we construct an alternative sequence-level split. Details about the split construction and the results are in Appendix~\ref{appendix: split info} and~\ref{app: split results}. We further evaluate our method on the occupancy prediction task following the experimental setting of RadarOcc~\cite{ding2024radarocc}.


\textbf{RADIal~\cite{Radial}} contains 91 driving sequences (about 2 hours in total) with around 8k
annotated frames, with radar signals released in raw ADC form. Since no
precomputed 4D radar cubes are provided, we reconstruct the 4D spectrum from the
raw ADC data using standard signal processing.
\subsection{Evaluation Protocol}
\label{sec:eval}
We adopt 3D object detection as the main downstream task, as it is the most widely studied setting in radar perception. Following the official K-Radar protocol~\cite{kradar}, we report \apbev/\apthd with an IoU threshold of 0.3. For RADIal, we report mAP with an IoU threshold of 0.5 due to known issues with the default evaluation metric\footnote{\url{https://github.com/valeoai/RADIal/issues/46}}.  For occupancy prediction, we evaluate foreground-background segmentation and report the mean Intersection over Union (mIoU), averaged over the two classes, at three evaluation ranges: 12.8\,m, 25.6\,m, and 51.2\,m.

To evaluate self-supervised learning methods, we consider two evaluation protocols. For fine-tuning, the encoder backbone is unfrozen and jointly optimized with the detection head to evaluate the transferability of the pretrained representations. The main results are discussed under the fine-tuning setting. Additionally, we conduct decoder probing, where the encoder is frozen while a lightweight two-layer decoder and the detection head are trained on top of it. Following recent work~\cite{nomae,wu2025sonata}, this protocol provides a more direct evaluation of the quality of the learned representation. We use it for the hyperparameter analysis (Figure~\ref{fig: ablations}) and the additional comparisons in Appendix~\ref{app: more results}.

\subsection{Implementation Details}
\label{sec: implement}
Following prior work~\cite{abdelsamad2026dos,wu2025sonata,nomae}, we adopt the
state-of-the-art 3D detector PTv3~\cite{ptv3} for all experiments. The voxel
size is set to $[0.4, 0.4, 0.4]$ m. The detection range of $[x,y,z]$ is
$[0, 72]\times[-6.4, 6.4]\times[-2, 6]$ m for K-Radar and
$[0, 103]\times[-40, 40]\times[-2.5, 2.0]$ m for RADIal. The encoder and decoder depths are set to $[2, 6]$ and $[4]$, respectively.
During pretraining, models are trained for 60k iterations with a learning rate
of $1\times10^{-3}$. For fine-tuning, we follow~\cite{nomae} and apply
layer-wise learning rate decay (LLRD)~\cite{he2022mae}, using a base learning
rate of $1\times10^{-4}$ for 20 epochs with a batch size of 8. For SPMAE, the spatial masking ratio is $m_{\mathrm{rate}}=0.7$. For L-SPMAE,
$m_{\mathrm{rate}}=0.4$, and spectral leakage is applied with $l_{\mathrm{rate}}=0.2$ along the
3D axes using a block size of $b=4$. We use CenterHead~\cite{yin2021center} with a score threshold of 0.1.
Additional details are provided in Appendix~\ref{app: hyper}.

\section{Results and Analysis}
\label{sec: results}

\subsection{From Handcrafted Reductions to Signal Discovery}
\begin{table}[t]
\centering
\caption{Comparison of \apbev/\apthd on the K-Radar official split.
Sup.\ denotes supervised learning. RAE is defined in
Section~\ref{sec: spmae}, and RAE-d and RAD in Section~\ref{sec: doppler_encoding}.
The best results are highlighted in \textbf{bold}.}
\footnotesize
\setlength{\tabcolsep}{8.5pt}

\begin{tabular}{clccccc}
\toprule
\multicolumn{2}{c}{Methods} & Representation & Sedan & Bus/Truck & Pedestrian & mAP \\
\midrule

\multirow{3}{*}{\rotatebox[origin=c]{90}{Sup.}} 
 & \multirow{3}{*}{PTv3~\cite{ptv3}} 
 & Point Cloud & 56.5/50.9 & 51.3/39.2 & 39.1/36.9 & 49.0/42.3 \\
 & & RAE   & 56.5/49.8 & 50.1/39.2 & 35.5/33.3 & 47.4/40.8 \\
 & & RAE-d & 53.4/47.3 & 46.2/36.8 & 37.7/34.5 & 45.8/39.5 \\

\midrule

\multirow{7}{*}{\rotatebox[origin=c]{90}{Fine-tune}}  
 & +OccupancyMAE~\cite{min2023occupancy} & Point Cloud & 58.6/53.0 & 54.0/45.5 & 38.5/37.6 & 50.4/45.4 \\  
 & +SONATA~\cite{wu2025sonata}        & Point Cloud & 58.2/52.6 & 51.4/41.0 & 42.7/41.6 & 50.8/45.0 \\  
 & +NOMAE~\cite{nomae}                & Point Cloud & 59.3/53.7 & 53.8/43.5 & 39.2/37.3 & 50.8/44.9 \\

\cmidrule(lr){2-7}

 & \textbf{+SPMAE (Ours)}   & RAE   & 61.0/55.0 & 55.7/48.1 & 42.5/41.2 & 53.1/48.1 \\
 & \bestcell{\textbf{+L-SPMAE (Ours)}} 
 & \bestcell{RAE} 
 & \bestcell{61.3/55.3} 
 & \bestcell{56.9/47.8} 
 & \bestcell{42.9/42.0} 
 & \bestcell{53.7/48.4} \\

 & \textbf{+SPMAE (Ours)}   & RAE-d & 59.3/53.4 & 54.2/44.8 & 42.9/40.8 & 52.1/46.3 \\
 & \textbf{+L-SPMAE (Ours)} & RAE-d & 59.3/53.4 & 54.9/45.7 & 42.4/41.4 & 52.2/46.8 \\

\midrule
\midrule

Sup.
 & PTv3~\cite{ptv3}          & RAD & 58.1/51.6 & 51.4/43.5 & 37.1/34.3 & 48.9/43.1 \\
\multirow{2}{*}{Fine-tune}
 & \textbf{+SPMAE (Ours)}   & RAD & 61.2/55.0 & 58.2/47.2 & 43.5/41.5 & 54.3/47.9 \\
 & \bestcell{\textbf{+L-SPMAE (Ours)}} 
 & \bestcell{RAD} 
 & \bestcell{\textbf{62.7/55.3}} 
 & \bestcell{\textbf{59.5/51.7}} 
 & \bestcell{\textbf{45.6/43.4}} 
 & \bestcell{\textbf{55.9/50.1}} \\

\bottomrule
\end{tabular}
\label{tab: split1}
\vspace{-0.3cm}
\end{table}

Table~\ref{tab: split1} reports results on the K-Radar benchmark.
Under fully supervised training, richer representations counterintuitively perform worse: compared with point clouds, RAE and RAE-d decrease performance by 1.6/1.5 and 3.2/2.8 mAP, respectively. This is consistent with the argument in
Section~\ref{sec: theory}: the supervised objective alone provides insufficient
guidance to interpret dense spectral structures, and the coherence structure latent
in the spectrum remains undiscovered.

Signal discovery reverses this trend. After fine-tuning, SPMAE and L-SPMAE
consistently outperform all point-cloud-based counterparts across all spectral
representations, with L-SPMAE on RAE reaching $53.7/48.4$ mAP ($+2.9/{+}3.0$ over
the best baseline). This confirms that the \mbox{pretraining} objective forces the encoder
to discover and represent coherent structures in the spectrum that handcrafted
reductions discard.

Comparing L-SPMAE and SPMAE, spectral leakage modeling consistently improves
performance across all representations ($+0.6/{+}0.3$ mAP on RAE), demonstrating
that physics-inspired corruption defines a harder and more structured prediction
task, yielding more informative representations. 

Despite containing more information than RAE, RAE-d underperforms after fine-tuning,
suggesting that stacking Doppler as channels prevents the model from
exploiting Doppler coherence effectively. The RAD representation addresses this by
exposing Doppler as a dedicated axis, preserving the same information as RAE-d but
in a layout that makes Doppler structure accessible to the model. L-SPMAE on RAD
achieves $\mathbf{55.9/50.1}$ mAP, surpassing all point-cloud baselines by +5.1/+4.7 points in \apbev/\apthd and recovering the expected ordering where richer representations yield
better performance after signal discovery. With the PTv3 backbone and RAD representation fixed, SPMAE improves over supervised training by +5.4/+4.8 \apbev/\apthd, while L-SPMAE provides a further gain of +1.6/+2.2 over SPMAE. These comparisons distinguish the benefit of masked pretraining from the additional contribution of leakage corruption.

\begin{table}[t]
\begin{minipage}{0.5\linewidth}
    \centering
    \captionof{table}{Comparison of \apbev/\apthd on the K-Radar benchmark: single-class ``Sedan'', label version v1.0, $\text{score\_threshold}=0.3$. Best results in \textbf{bold}.}
    \footnotesize
\setlength{\tabcolsep}{4pt}
\begin{tabular}{lcc}
\toprule
Methods&Representation& Sedan \\
\midrule
RTNH~\cite{kradar}&Point Cloud& 58.4/47.4 \\
DPFT~\cite{fent2024dpft}&Image+Reduced Spectrum&-/56.1\\
RTNH+~\cite{Kong2025RTNHplus} &Reduced Spectrum& 65.7/57.6 \\
PTv3~\cite{ptv3}&Spectrum& 61.9/58.4\\
\midrule
\textbf{+L-SPMAE(Ours)}&Spectrum& \textbf{67.7/64.0} \\
\bottomrule
\end{tabular}

    \label{tab: labelv1.0}
\end{minipage}
\hspace{20pt}
\begin{minipage}{0.4\linewidth}
    \centering
    \captionof{table}{Comparison of \apbev on the \mbox{RADIal} test set. We follow the same evaluation protocol as K-Radar, reporting mAP based on the PR curve with IoU=0.5.}
    \footnotesize
\begin{tabular}{lcc}
\toprule
Methods&Representation& mAP\\
\midrule
FFTRadNet~\cite{Radial}&RD map& 89.3 \\
PTv3~\cite{ptv3}&Spectrum&89.0\\
\midrule
\textbf{+SPMAE (Ours)} &Spectrum& 89.6\\
\textbf{+L-SPMAE (Ours)}&Spectrum&\textbf{90.0}\\
\bottomrule
\end{tabular}

    \label{tab: radial}
\end{minipage}
    \vspace{-0.5cm}
\end{table}

\subsection{Comparison with State-of-the-Art Radar Detectors}
Existing radar detectors typically rely on handcrafted reduced representations
with radar-specific architectural designs, as discussed in Section~\ref{sec:
related}. We investigate whether signal discovery on raw radar spectra using a
generic 3D backbone can learn more representative features. Following the
official K-Radar benchmarking protocol, Table~\ref{tab: labelv1.0} shows that
without pretraining, PTv3 already achieves higher \apthd than RTNH+~\cite{Kong2025RTNHplus}, and
with fine-tuning further obtains gains of $+2.0/{+}6.4$ in \apbev/\apthd over RTNH+.

Table~\ref{tab: 4class compare} reports results under the 4-class setting of
RADENet~\cite{leitgeb2026rade}. PTv3 with RAD already surpasses RADENet across
all classes without pretraining, with particularly strong gains on Pedestrian
and Bicycle. Fine-tuning further amplifies these gains, achieving $+6.6/{+}7.0$
on the Sedan class, and L-SPMAE consistently outperforms SPMAE by $+1.5/{+}1.4$.
Performance degradation on Bicycle after fine-tuning is attributed to the
extremely limited number of training samples for that class.

\subsection{Generalization to Other Datasets and Tasks}
\label{sec: generalization}
Table~\ref{tab: radial} reports results on RADIal, where spectrum inputs are
constructed from raw radar signals. L-SPMAE achieves the best performance
($+0.7$ over FFTRadNet), though gains are smaller than on K-Radar, which we
attribute to the lower scene diversity and complexity of RADIal.

We further evaluate our methods on radar-based occupancy prediction, with the results summarized in Table~\ref{tab:occupancy_results}. Both SPMAE and L-SPMAE consistently improve over the supervised PTv3 baseline across all evaluation ranges. Compared with RadarOcc, our models achieve substantially higher foreground IoU, while RadarOcc performs better on background IoU, indicating that our pretrained representations are particularly effective at capturing foreground objects. The gain of L-SPMAE over SPMAE is relatively modest, likely because the binary foreground-background formulation provides only coarse supervision. We additionally evaluate the learned representations on radar-based multi-object tracking, with results provided in Appendix~\ref{sec: Tracking}.
\begin{table}[t]
\centering
\caption{Occupancy prediction results on K-Radar following RadarOcc~\cite{ding2024radarocc}. Reproduced RadarOcc results slightly outperform the reported values.}
\label{tab:occupancy_results}
\small
\footnotesize
\begin{tabular}{lccccccccc}
\toprule
\multirow{2}{*}{Methods}
& \multicolumn{3}{c}{mIoU (\%)} 
& \multicolumn{3}{c}{BG IoU (\%)} 
& \multicolumn{3}{c}{FG IoU (\%)} \\
\cmidrule(lr){2-4}
\cmidrule(lr){5-7}
\cmidrule(lr){8-10}
& 12.8m & 25.6m & 51.2m
& 12.8m & 25.6m & 51.2m
& 12.8m & 25.6m & 51.2m \\
\midrule

RadarOcc~\cite{ding2024radarocc}
& 42.1 & 33.7 & 25.7
& \textbf{49.7} & \textbf{41.0} & \textbf{30.6}
& 34.5 & 26.4 & 20.8 \\

PTv3~\cite{ptv3}
& 45.4 & 34.2 & 26.8
& 48.2 & 36.6 & 28.4
& 42.5 & 31.8 & 25.2 \\

\midrule

\textbf{+SPMAE (Ours)}
& 47.2 & 35.1 & 27.4
& 49.0 & 37.6 & 28.8
& 45.3 & 32.6 & 25.9 \\

\textbf{+L-SPMAE (Ours)}
& \textbf{47.5} & \textbf{35.5} & \textbf{27.8}
& 49.3 & 37.7 & 29.1
& \textbf{45.8} & \textbf{33.3} & \textbf{26.5} \\

\bottomrule
\end{tabular}
\end{table}



\begin{table}[t]
    \vspace{-7pt}
    \caption{\small Comparison of \apbev/\apthd on the K-Radar official split with RADENet 4-class setting, label version v2.1.}
  \centering
\footnotesize
\begin{tabular}{lcccccc}
\toprule
Methods & Representation & Sedan & Bus/Truck & Pedestrian & Bicycle & mAP \\
\midrule
 RADENet~\cite{leitgeb2026rade} 
& Reduced Spectrum 
& 63.2/56.8 & 48.3/41.0 & 6.8/5.4 & 1.5/1.5 & 30.0/26.2 \\
PTv3~\cite{ptv3} 
& Spectrum 
& 64.7/58.8 & 61.6/50.8 & 35.6/34.4& \textbf{49.0/42.2}& 52.7/46.5\\
\midrule

\textbf{+SPMAE (Ours)} 
& Spectrum 
&67.8/61.8 & \textbf{65.1/53.2} & 48.7/47.0 & 29.7/29.7 &52.8/47.9 \\

 \textbf{+L-SPMAE (Ours)} 
& Spectrum 
& \textbf{69.8/63.8} & 63.9/52.1 & \textbf{49.9/47.6} & 33.6/33.6 & \textbf{54.3/49.3} \\
\bottomrule
\end{tabular}

  \label{tab: 4class compare}
  \vspace{-14pt}
\end{table}

\subsection{Ablation Study}
\label{sec: ablation}
Ablation studies are conducted on the K-Radar official split for three classes with label version v2.0. Additional results and visualizations are provided in Appendix~\ref{app: more results}.

\textbf{Variants of L-SPMAE.}
Table~\ref{tab:variant} compares different masking and corruption strategies. The FOCUS masking~\cite{fang2024prism} improves over the supervised baseline, but remains below global random masking (SPMAE). This highlights a key distinction: FOCUS relies on predefined power-based selection of informative regions, whereas SPMAE/L-SPMAE operate on the complete unreduced spectrum without such preselection. Adding leakage corruption to FOCUS yields only a modest gain, while applying leakage globally performs substantially better. Combining global random masking with leakage modeling achieves the best performance, suggesting that full-spectrum learning and leakage-aware corruption provide complementary benefits.

\begin{table}[tb]
\centering

\begin{minipage}[t]{0.48\linewidth}
    \centering
    \captionof{table}{Comparison of variants. FOCUS Masking is reimplemented following the original paper.}
    \label{tab:variant}

\footnotesize
\begin{tabular}{lcc}
\toprule
Method & \apbev & \apthd \\
\midrule
From Scratch & 48.9 & 43.1 \\
\midrule
 FOCUS Mask~\cite{fang2024prism} & 52.1 & 45.4 \\
FOCUS Mask + Leakage & 52.4 & 45.7 \\
Leakage Only & 53.9 & 47.8 \\
 Random Mask  & 54.3 & 47.9 \\
\textbf{Leakage + Random Mask}
& \textbf{55.9} & \textbf{50.1} \\
\bottomrule
\end{tabular}
\end{minipage}
\hfill
\begin{minipage}[t]{0.48\linewidth}
    \centering
    \captionof{table}{Comparison of radar-only L-SPMAE with LiDAR-guided pretraining approaches using the same PTv3 backbone on the K-Radar official split.}
    \label{tab:lidar_guided}
    \footnotesize
\begin{tabular}{lc}
\toprule
Methods & mAP \\
\midrule
PTv3~\cite{ptv3} &48.9 / 43.1\\
\midrule
+ LiDAR Occ Pretrain~\cite{huang2025towards} & 52.2 / 45.9 \\
+ LiDAR Feat Distill~\cite{bang2024radardistill} & 53.1 / 46.8 \\
\textbf{+ L-SPMAE (Ours)} & \textbf{55.9 / 50.1} \\
\bottomrule
\end{tabular}
\end{minipage}


\begin{minipage}[t]{0.48\linewidth}
    \centering
    \captionof{table}{ Results with the Sparse Encoder backbone.}
    \label{tab:sparse_encoder}
    \footnotesize
\begin{tabular}{lcc}
\toprule
 Backbone & \apbev & \apthd \\
\midrule
Sparse Encoder~\cite{yin2021center} & 47.2 & 40.5 \\
\midrule
 \textbf{+ SPMAE (Ours)} & 52.4 & 44.9 \\
\textbf{+ L-SPMAE (Ours)} & \textbf{53.1} & \textbf{47.0} \\
\bottomrule
\end{tabular}

\end{minipage}
\hfill
\begin{minipage}[t]{0.48\linewidth}
    \centering
    \captionof{table}{Effect of L-SPMAE reconstruction. A PTv3 is trained from scratch using the reconstructed spectra.}
    \vspace{4pt}
    \label{tab:reconstructed_spectra}
    \footnotesize
\begin{tabular}{lcc}
\toprule
Representation & \apbev & \apthd \\
\midrule
PTv3+Original Spectra      & 48.9 & 43.1 \\
PTv3+Reconstructed Spectra & \textbf{52.0} & \textbf{45.7} \\
\bottomrule
\end{tabular}
\end{minipage}
\vspace{-0.5cm}
\end{table}
\textbf{Backbone Choices.}
To verify that our method is not backbone-specific, we additionally report results with a Sparse Encoder~\citep{yin2021center} in Table~\ref{tab:sparse_encoder}. SPMAE clearly improves over the supervised baseline, and L-SPMAE brings further gains, showing the same trend as with PTv3. These results confirm that our leakage-inspired pretraining generalizes across backbone architectures.

\textbf{Comparison with LiDAR-Guided Pretraining.}
Recent methods~\citep{bang2024radardistill, huang2025towards} use LiDAR as auxiliary supervision to provide cleaner geometry for radar perception. We compare our radar-only pretraining with two LiDAR-guided approaches in Table~\ref{tab:lidar_guided}. L-SPMAE clearly outperforms LiDAR feature distillation and LiDAR occupancy pretraining. This suggests that constraining radar representations toward clean LiDAR geometry may discard useful radar-specific context, while L-SPMAE better preserves and exploits such task-relevant structures.

\textbf{Empirical Validation of Spectral Reconstruction.}
To empirically validate the analysis in Section~\ref{sec: theory}, we reconstruct each radar spectrum in a single unmasked forward pass through a frozen L-SPMAE, then train a new PTv3 from scratch using only the reconstructed spectra. As shown in Table~\ref{tab:reconstructed_spectra}, the reconstructed
spectra improve \apbev from 48.9 to 52.0 and \apthd
from 43.1 to 45.7. Since the downstream network is trained from scratch, this
gain is independent of weight initialization, suggesting that L-SPMAE suppresses
task-irrelevant variations while preserving coherent structures useful for perception.

\begin{figure}[h]
    \centering
    \begin{subfigure}{0.32\linewidth}
        \centering
        \includegraphics[width=\linewidth]{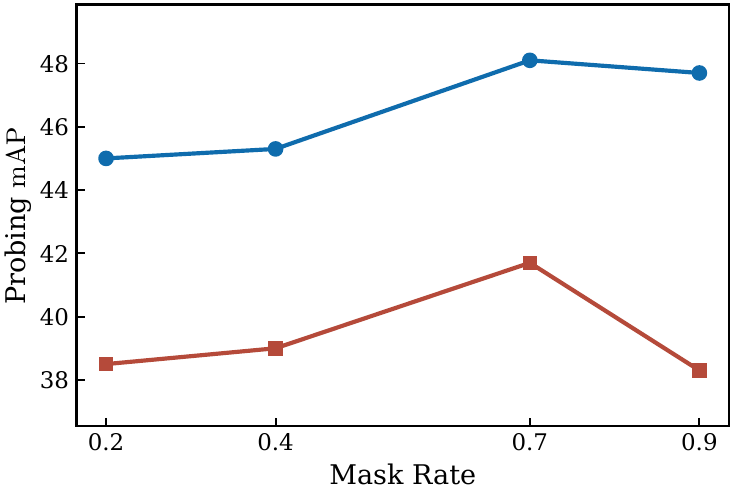}
        \caption{}
        \label{fig: mask_rate}
    \end{subfigure}
    \hfill
    \begin{subfigure}{0.32\linewidth}
        \centering
        \includegraphics[width=\linewidth]{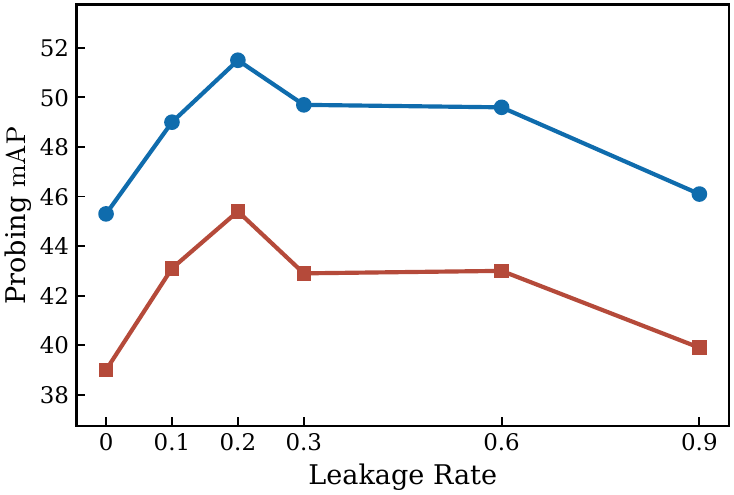}
        \caption{}
        \label{fig: leakage_rate}
    \end{subfigure}
    \hfill
    \begin{subfigure}{0.32\linewidth}
        \centering
        \includegraphics[width=\linewidth]{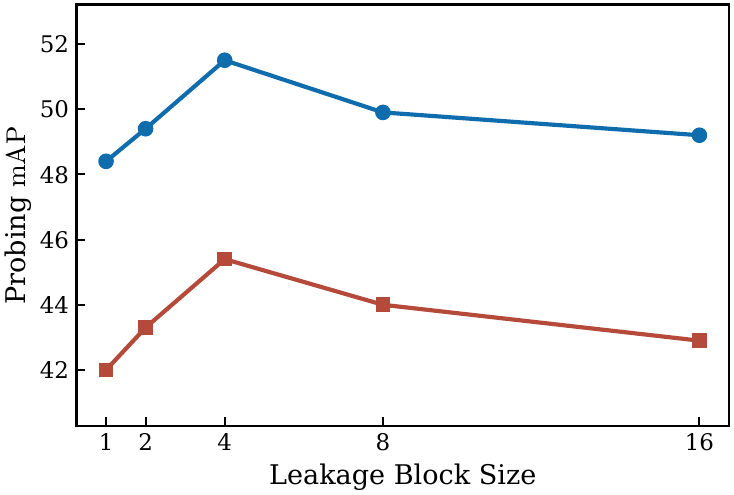}
        \caption{}
        \label{fig: block size}
    \end{subfigure}
    \caption{Probing results of hyperparameter settings: (a) mask rate $m_{\text{rate}}$ (SPMAE, no leakage), (b) leakage rate $l_{\text{rate}}$, and (c) leakage block size $b$. (\textcolor{APBEV}{$AP_{\text{BEV}}$}/\textcolor{APTHREED}{$AP_{\text{3D}}$}).}
    
    \label{fig: ablations}
    \vspace{-0.5cm}
\end{figure}

\textbf{Hyperparameter Analysis.}
Figure~\ref{fig: ablations} studies the impact of the masking rate $m_{\text{rate}}$,
leakage rate $l_{\text{rate}}$, and leakage block size $b$. The masking rate
controls the difficulty of the pretext task, where $m_{\text{rate}}=0.7$ performs
best. The leakage rate determines the amount of energy spread: values in
$0.1$--$0.6$ are robust with $l_{\text{rate}}=0.2$ performing best, while
overly aggressive leakage (e.g., $0.9$) degrades performance. The block size $b$
controls the locality of leakage, with $b=4$ yielding the best overall performance.

\vspace{-0.3cm}

\section{Conclusion}
\label{sec: conclusion}
We have argued and empirically demonstrated that handcrafted reductions of the
radar spectrum such as point clouds discard coherent structure that supervised
training cannot recover, but that corrupted-context prediction can discover.
Under fully supervised training, richer spectral representations perform worse
than point clouds; after signal discovery pretraining, this ordering reverses,
with L-SPMAE on the RAD representation surpassing all point-cloud-based
counterparts by a substantial margin. This reversal is the central finding of
our work: the information is present in the spectrum, but requires the right
pretraining objective to be unlocked. We further show that physics-inspired
corruption via spectral leakage modeling defines a harder and more structured
prediction task than random masking alone, and that the choice of representation
layout, not merely the quantity of retained information, determines how
effectively the coherence structure of the spectrum can be exploited.

\textbf{Limitations and Future Work}. 
Processing dense radar spectra incurs substantially higher computational cost than point-cloud-based perception. Future work will explore token pruning ~\cite{gitmerge3d2025} to reduce this overhead, as well as alternative SSL methods beyond MAE for spectrum representations.
\newpage
\section*{Acknowledgements and Disclosures}
This work was supported in part by the IPCEI ME/CT project. The IPCEI ME/CT project is supported by the Federal Ministry for Economic Affairs and Climate Action on the basis of a decision by the German Parliament, by the Ministry for Economic Affairs, Labor and Tourism of Baden-Württemberg based on a decision of the State Parliament of Baden-Württemberg, the Free State of Saxony on the basis of the budget adopted by the Saxon State Parliament, the Bavarian State Ministry for Economic Affairs, Regional Development and Energy and financed by the European Union – NextGenerationEU.
\bibliographystyle{unsrt}
\bibliography{ref}

\begin{thebibliography}{10}

\bibitem{schramm2024bevcar}
Jonas Schramm, Niclas V{\"o}disch, K{\"u}rsat Petek, B~Ravi Kiran, Senthil Yogamani, Wolfram Burgard, and Abhinav Valada.
\newblock Bevcar: Camera-radar fusion for bev map and object segmentation.
\newblock In {\em 2024 IEEE/RSJ International Conference on Intelligent Robots and Systems (IROS)}, pages 1435--1442. IEEE, 2024.

\bibitem{rohling2007radar}
Hermann Rohling.
\newblock Radar cfar thresholding in clutter and multiple target situations.
\newblock {\em IEEE Transactions on Aerospace and Electronic Systems}, AES-19(4):608--621, 1983.

\bibitem{palffy2022vod}
Andras Palffy, Ewoud Pool, Srimannarayana Baratam, Julian~FP Kooij, and Dariu~M Gavrila.
\newblock Multi-class road user detection with 3+ 1d radar in the view-of-delft dataset.
\newblock {\em IEEE Robotics and Automation Letters}, 7(2):4961--4968, 2022.

\bibitem{liu2023smurf}
Jianan Liu, Qiuchi Zhao, Weiyi Xiong, Tao Huang, Qing-Long Han, and Bing Zhu.
\newblock Smurf: Spatial multi-representation fusion for 3d object detection with 4d imaging radar.
\newblock {\em IEEE Transactions on Intelligent Vehicles}, 9(1):799--812, 2023.

\bibitem{ding2024radarocc}
Fangqiang Ding, Xiangyu Wen, Yunzhou Zhu, Yiming Li, and Chris~Xiaoxuan Lu.
\newblock Radarocc: Robust 3d occupancy prediction with 4d imaging radar.
\newblock {\em Advances in Neural Information Processing Systems}, 37:101589--101617, 2024.

\bibitem{leitgeb2026rade}
Christof Leitgeb, Thomas Puchleitner, Max~Peter Ronecker, and Daniel Watzenig.
\newblock Rade-net: Robust attention network for radar-only object detection in adverse weather.
\newblock {\em arXiv preprint arXiv:2602.19994}, 2026.

\bibitem{parit2026real}
Santosh Parit, Julia Hindel, Rohit Mohan, Georgi Pramatarov, Paul Newman, Daniele De~Martini, and Abhinav Valada.
\newblock Real-time radar semantic and panoptic scene understanding.
\newblock {\em arXiv preprint arXiv:2610.26816}, 2026.

\bibitem{nayak2024ralf}
Abhijeet Nayak, Daniele Cattaneo, and Abhinav Valada.
\newblock Ralf: Flow-based global and metric radar localization in lidar maps.
\newblock In {\em 2024 IEEE International Conference on Robotics and Automation (ICRA)}, pages 5097--5103. IEEE, 2024.

\bibitem{cheng2024centerradarnet}
Jen-Hao Cheng, Sheng-Yao Kuan, Hou-I Liu, Hugo Latapie, Gaowen Liu, and Jenq-Neng Hwang.
\newblock Centerradarnet: Joint 3d object detection and tracking framework using 4d fmcw radar.
\newblock In {\em 2024 IEEE International Conference on Image Processing (ICIP)}, pages 998--1004. IEEE, 2024.

\bibitem{Kong2025RTNHplus}
Seung-Hyun Kong, Dong-Hee Paek, and Sangyeong Lee.
\newblock Rtnh+: Enhanced 4d radar object detection network using two-level preprocessing and vertical encoding.
\newblock {\em IEEE Transactions on Intelligent Vehicles}, 10:1427--1440, 2025.

\bibitem{peng2024mufasa}
Xiangyuan Peng, Miao Tang, Huawei Sun, Kay Bierzynski, Lorenzo Servadei, and Robert Wille.
\newblock Mufasa: Multi-view fusion and adaptation network with spatial awareness for radar object detection.
\newblock In {\em International Conference on Artificial Neural Networks}, pages 168--184. Springer, 2024.

\bibitem{paek2023enhanced}
Dong-Hee Paek, Seung-Hyun Kong, and Kevin~Tirta Wijaya.
\newblock Enhanced k-radar: Optimal density reduction to improve detection performance and accessibility of 4d radar tensor-based object detection.
\newblock In {\em 2023 IEEE Intelligent Vehicles Symposium (IV)}, pages 1--6. IEEE, 2023.

\bibitem{batson2019noise2self}
Joshua Batson and Loic Royer.
\newblock Noise2self: Blind denoising by self-supervision.
\newblock In {\em International Conference On Machine Learning}, pages 524--533. PMLR, 2019.

\bibitem{kradar}
Dong-Hee Paek, Seung-Hyun Kong, and Kevin~Tirta Wijaya.
\newblock K-radar: 4d radar object detection for autonomous driving in various weather conditions.
\newblock {\em Advances in Neural Information Processing Systems}, 35:3819--3829, 2022.

\bibitem{ptv3}
Xiaoyang Wu, Li~Jiang, Peng-Shuai Wang, Zhijian Liu, Xihui Liu, Yu~Qiao, Wanli Ouyang, Tong He, and Hengshuang Zhao.
\newblock Point transformer v3: Simpler faster stronger.
\newblock In {\em Proceedings of the IEEE/CVF conference on computer vision and pattern recognition}, pages 4840--4851, 2024.

\bibitem{Radial}
Julien Rebut, Arthur Ouaknine, Waqas Malik, and Patrick P\'erez.
\newblock Raw high-definition radar for multi-task learning.
\newblock In {\em Proceedings of the IEEE/CVF Conference on Computer Vision and Pattern Recognition (CVPR)}, pages 17021--17030, June 2022.

\bibitem{han20234d}
Zeyu Han, Jiahao Wang, Zikun Xu, Shuocheng Yang, Lei He, Shaobing Xu, Jianqiang Wang, and Keqiang Li.
\newblock 4d millimeter-wave radar in autonomous driving: A survey.
\newblock {\em arXiv preprint arXiv:2306.04242}, 2023.

\bibitem{wang2025sddiff}
Shengpeng Wang, Xin Luo, Yulong Xie, and Wei Wang.
\newblock Sddiff: Boosting radar perception via spatial-doppler diffusion.
\newblock In {\em Proceedings of the Thirty-Fourth International Joint Conference on Artificial Intelligence}, pages 8805--8813, 2025.

\bibitem{roldan2024see}
Ignacio Roldan, Andras Palffy, Julian~FP Kooij, Dariu~M Gavrila, Francesco Fioranelli, and Alexander Yarovoy.
\newblock See further than cfar: a data-driven radar detector trained by lidar.
\newblock In {\em 2024 IEEE Radar Conference (RadarConf24)}, pages 1--6. IEEE, 2024.

\bibitem{zeller2023rit}
Matthias Zeller, Vardeep~S Sandhu, Benedikt Mersch, Jens Behley, Michael Heidingsfeld, and Cyrill Stachniss.
\newblock Radar instance transformer: Reliable moving instance segmentation in sparse radar point clouds.
\newblock {\em IEEE Transactions on Robotics}, 40:2357--2372, 2023.

\bibitem{tian2023geomae}
Xiaoyu Tian, Haoxi Ran, Yue Wang, and Hang Zhao.
\newblock Geomae: Masked geometric target prediction for self-supervised point cloud pre-training.
\newblock In {\em Proceedings of the IEEE/CVF Conference on Computer Vision and Pattern Recognition}, pages 13570--13580, 2023.

\bibitem{min2023occupancy}
Chen Min, Liang Xiao, Dawei Zhao, Yiming Nie, and Bin Dai.
\newblock Occupancy-mae: Self-supervised pre-training large-scale lidar point clouds with masked occupancy autoencoders.
\newblock {\em IEEE Transactions on Intelligent Vehicles}, 9(7):5150--5162, 2023.

\bibitem{nomae}
Mohamed Abdelsamad, Michael Ulrich, Claudius Gl{\"a}ser, and Abhinav Valada.
\newblock Multi-scale neighborhood occupancy masked autoencoder for self-supervised learning in lidar point clouds.
\newblock In {\em Proceedings of the Computer Vision and Pattern Recognition Conference}, pages 22234--22243, 2025.

\bibitem{wu2025sonata}
Xiaoyang Wu, Daniel DeTone, Duncan Frost, Tianwei Shen, Chris Xie, Nan Yang, Jakob Engel, Richard Newcombe, Hengshuang Zhao, and Julian Straub.
\newblock Sonata: Self-supervised learning of reliable point representations.
\newblock In {\em Proceedings of the Computer Vision and Pattern Recognition Conference}, pages 22193--22204, 2025.

\bibitem{abdelsamad2026dos}
Mohamed Abdelsamad, Michael Ulrich, Bin Yang, Miao Zhang, Yakov Miron, and Abhinav Valada.
\newblock Dos: Distilling observable softmaps of zipfian prototypes for self-supervised point representation.
\newblock In {\em Proceedings of the AAAI Conference on Artificial Intelligence}, volume~40, pages 19514--19523, 2026.

\bibitem{yang2026pointins}
Bin Yang, Mohamed Abdelsamad, Miao Zhang, and Alexandru~Paul Condurache.
\newblock Towards foundation models for 3d scene understanding: Instance-aware self-supervised learning for point clouds.
\newblock In {\em Proceedings of the IEEE/CVF Conference on Computer Vision and Pattern Recognition (CVPR)}, pages 2938--2947, June 2026.

\bibitem{abdelsamad2026ghostpoint}
Mohamed Abdelsamad, Bin Yang, Michael Ulrich, Miao Zhang, Yakov Miron, Alexandru~Paul Condurache, and Abhinav Valada.
\newblock Ghostpoint: Self-supervised representation learning by hallucinating occluded lidar structure.
\newblock In {\em European Conference on Computer Vision}, pages 116--137. Springer, 2026.

\bibitem{ding2022raflow}
Fangqiang Ding, Zhijun Pan, Yimin Deng, Jianning Deng, and Chris~Xiaoxuan Lu.
\newblock Self-supervised scene flow estimation with 4-d automotive radar.
\newblock {\em IEEE Robotics and Automation Letters}, 7(3):8233--8240, 2022.

\bibitem{li2025super4dr}
Zhiheng Li, Weihua Wang, Qiang Shen, Yichen Zhao, and Zheng Fang.
\newblock Super4dr: 4d radar-centric self-supervised odometry and gaussian-based map optimization.
\newblock {\em arXiv preprint arXiv:2512.09608}, 2025.

\bibitem{hao2024bootstrapping}
Yiduo Hao, Sohrab Madani, Junfeng Guan, Mohammed Alloulah, Saurabh Gupta, and Haitham Hassanieh.
\newblock Bootstrapping autonomous driving radars with self-supervised learning.
\newblock In {\em Proceedings of the IEEE/CVF Conference on Computer Vision and Pattern Recognition}, pages 15012--15023, 2024.

\bibitem{luo2025radler}
Yuan Luo, Rudolf Hoffmann, Yan Xia, Olaf Wysocki, Benedikt Schwab, Thomas~H Kolbe, and Daniel Cremers.
\newblock Radler: Radar object detection leveraging semantic 3d city models and self-supervised radar-image learning.
\newblock In {\em Proceedings of the Computer Vision and Pattern Recognition Conference}, pages 4452--4461, 2025.

\bibitem{fang2024prism}
Liang Fang, Ruiyuan Song, Zhi Lu, Dongheng Zhang, Yang Hu, Qibin Sun, and Yan Chen.
\newblock Prism: Pre-training rf signals in sparsity-aware masked autoencoders.
\newblock In {\em IEEE INFOCOM 2024-IEEE Conference on Computer Communications}, pages 2109--2118. IEEE, 2024.

\bibitem{lyon2009discrete}
Douglas~A Lyon.
\newblock The discrete fourier transform, part 4: spectral leakage.
\newblock {\em Journal of object technology}, 8(7), 2009.

\bibitem{fent2024dpft}
Felix Fent, Andras Palffy, and Holger Caesar.
\newblock Dpft: Dual perspective fusion transformer for camera-radar-based object detection.
\newblock {\em IEEE Transactions on Intelligent Vehicles}, 2024.

\bibitem{zhang2024exploring}
Miao Zhang, Sherif Abdulatif, Benedikt Loesch, Marco Altmann, Marius Schwarz, and Bin Yang.
\newblock Exploring domain shift on radar-based 3d object detection amidst diverse environmental conditions.
\newblock In {\em 2024 IEEE 27th International Conference on Intelligent Transportation Systems (ITSC)}, pages 3513--3518. IEEE, 2024.

\bibitem{Zhang2025capmix}
Miao Zhang, Sherif Abdulatif, Benedikt Loesch, Marco Altmann, and Bin Yang.
\newblock Class-aware pillarmix: Can mixed sample data augmentation enhance 3d object detection with radar point clouds?
\newblock {\em 2025 IEEE/RSJ International Conference on Intelligent Robots and Systems (IROS)}, pages 13349--13356, 2025.

\bibitem{he2022mae}
Kaiming He, Xinlei Chen, Saining Xie, Yanghao Li, Piotr Doll{\'a}r, and Ross Girshick.
\newblock Masked autoencoders are scalable vision learners.
\newblock In {\em Proceedings of the IEEE/CVF conference on computer vision and pattern recognition}, pages 16000--16009, 2022.

\bibitem{yin2021center}
Tianwei Yin, Xingyi Zhou, and Philipp Krahenbuhl.
\newblock Center-based 3d object detection and tracking.
\newblock In {\em Proceedings of the IEEE/CVF conference on computer vision and pattern recognition}, pages 11784--11793, 2021.

\bibitem{huang2025towards}
Tianshu Huang, Akarsh Prabhakara, Chuhan Chen, Jay Karhade, Deva Ramanan, Matthew O'toole, and Anthony Rowe.
\newblock Towards foundational models for single-chip radar.
\newblock In {\em 2025 IEEE/CVF International Conference on Computer Vision (ICCV)}, pages 24655--24665. IEEE, 2025.

\bibitem{bang2024radardistill}
Geonho Bang, Kwangjin Choi, Jisong Kim, Dongsuk Kum, and Jun~Won Choi.
\newblock Radardistill: Boosting radar-based object detection performance via knowledge distillation from lidar features.
\newblock In {\em 2024 IEEE/CVF Conference on Computer Vision and Pattern Recognition (CVPR)}, pages 15491--15500. IEEE, 2024.

\bibitem{gitmerge3d2025}
Tuan~Anh Tran, Duy Minh~Ho Nguyen, Hoai-Chau Tran, Michael Barz, Khoa~D. Doan, Roger Wattenhofer, Vien~Anh Ngo, Mathias Niepert, Daniel Sonntag, and Paul Swoboda.
\newblock How many tokens do 3d point cloud transformer architectures really need?
\newblock In {\em Advances in Neural Information Processing Systems}, 2025.

\bibitem{pang2022simpletrack}
Ziqi Pang, Zhichao Li, and Naiyan Wang.
\newblock Simpletrack: Understanding and rethinking 3d multi-object tracking.
\newblock In {\em European conference on computer vision}, pages 680--696. Springer, 2022.

\bibitem{yang2026collis}
Bin Yang and Alexandru~Paul Condurache.
\newblock Collaborative learning for semi-supervised lidar semantic segmentation.
\newblock In {\em International Conference on Machine Learning}, 2026.

\bibitem{pointcept2023}
Pointcept Contributors.
\newblock Pointcept: A codebase for point cloud perception research.
\newblock \url{https://github.com/Pointcept/Pointcept}, 2023.

\bibitem{mmdet3d2020}
MMDetection3D Contributors.
\newblock {MMDetection3D: OpenMMLab} next-generation platform for general {3D} object detection.
\newblock \url{https://github.com/open-mmlab/mmdetection3d}, 2020.

\bibitem{openpcdet2020}
OpenPCDet~Development Team.
\newblock Openpcdet: An open-source toolbox for 3d object detection from point clouds.
\newblock \url{https://github.com/open-mmlab/OpenPCDet}, 2020.

\end{thebibliography}
\newpage
\appendix
\section*{Appendix}
\section{Experimental Settings}
\subsection{Sequence Split Details of K-Radar}
\label{appendix: split info}
To enforce a stricter separation between training and testing data, we propose a new sequence split,
which is more challenging due to the limited number of sequences and the resulting domain shift. 
Table~\ref{tab: split_info} summarizes the proposed sequence-level split. We partition the dataset by sequence ID with a 75/25 train/test ratio, ensuring that each weather condition is represented in both splits. The reported numbers of frames and the annotation distribution are computed after applying our predefined ROI filtering.
\begin{table}[h]
\centering
\small
\setlength{\tabcolsep}{5pt}        
\renewcommand{\arraystretch}{1.15} 
\caption{Split statistics of the sequence-level split.}
\begin{tabularx}{\linewidth}{l >{\RaggedRight\arraybackslash}X >{\RaggedRight\arraybackslash}X}
\toprule
\textbf{Split Info} & \textbf{Train} & \textbf{Test} \\
\midrule
\textbf{Sequence IDs} &
1, 2, 3, 5, 6, 7, 8, 9, 10, 11, 12, 14, 16, 18, 19, 21, 22, 24, 25, 27, 28, 30, 31, 32, 33, 34, 35, 37, 38, 39, 40, 42, 43, 45, 46, 47, 48, 50, 51, 52, 54, 55, 56, 57 &
4, 13, 15, 17, 20, 23, 26, 29, 36, 41, 44, 49, 53, 58 \\
\textbf{Filtered frames} & 19{,}843 & 5{,}903 \\
\textbf{Filtered class distribution} &
\makecell[l]{Sedan: 45{,}251\\ Bus/Truck: 8{,}481\\ Pedestrian: 3{,}513} &
\makecell[l]{Sedan: 10{,}409\\ Bus/Truck: 2{,}483\\ Pedestrian: 1{,}408} \\
\bottomrule
\end{tabularx}
\label{tab: split_info}
\vspace{-15pt}
\end{table}

\subsection{Hyperparameter Settings}
\label{app: hyper}
\begin{table}[tb]
\centering

\begin{minipage}{0.48\linewidth}
\centering
\caption{PTv3 model setting.}
\resizebox{\linewidth}{!}{
\begin{tabular}{lc}
\toprule
Config & Value \\
\midrule
serialization pattern& Z+TZ+H+TH\\
patch interaction  & Shift Order+Shuffle Order\\
encoder depth&[2,6]\\
encoder channels&[64,128]\\
encoder num heads&[4,8]\\
encoder patch size&[128,128]\\
decoder depth&[4]\\
decoder channels&[64]\\
decoder num heads&[8]\\
decoder patch size&[128]\\
mlp ratio& [4]\\
\bottomrule
\end{tabular}
}
\label{tab: ptv3}
\end{minipage}
\hfill
\begin{minipage}{0.48\linewidth}
\centering
\caption{Settings of point-cloud-based SSL.}
\resizebox{\linewidth}{!}{
\begin{tabular}{lc}
\toprule
Methods& Settings \\
\midrule
OccupancyMAE~\cite{min2023occupancy} &
\makecell[c]{
$m_{\mathrm{rate}}$ in 0--30\,m: 0.7\\
$m_{\mathrm{rate}}$ in 30--50\,m: 0.5\\
$m_{\mathrm{rate}}$ in $>50$\,m: 0.3
}\\
\midrule
NOMAE~\cite{nomae}&\makecell[c]{neighborhood size $n=9$\\
$m_{\mathrm{rate}}=0.7$}\\
\midrule
SONATA~\cite{wu2025sonata}&\makecell[c]{ mask view:2\\
local view:4\\
$m_{\mathrm{rate}}:0.4 \rightarrow 0.7$\\
Augmentations: GlobalRotScaleTrans,\\
Flip3D, JitterPoints\\
}\\
\bottomrule
\end{tabular}
}

\label{tab: settings}
\end{minipage}

\end{table}
Table~\ref{tab: ptv3} summarizes the PTv3~\cite{ptv3} configuration in our work. Compared to the original setting for LiDAR and RGB-D camera, we adopt fewer stages and a shallower network.

We give the hyperparameter setting of point-cloud-based SSL in Table~\ref{tab: settings}. For OccupancyMAE~\cite{min2023occupancy}, we keep the range-wise masking but reduce the mask rate $m_{\mathrm{rate}}$ by 0.2, since the default setting is too aggressive for radar’s lower-resolution signals. For NOMAE~\cite{nomae}, all the default settings are maintained. For SONATA~\cite{wu2025sonata}, we limit rotation augmentation to a small range due to our front-view setup. Accordingly, BEV flipping is applied only horizontally.

\section{Additional Experiments}
\label{app: more results}
\subsection{Sequence Split Results}
\label{app: split results}
\begin{table}[t]
\centering
\caption{Comparison of \apbev/\apthd on the K-Radar sequence split.
Sup.\ denotes supervised learning. RAE is introduced in
Section~\ref{sec: spmae}, and RAD in Section~\ref{sec: doppler_encoding}.
The best results are highlighted in \textbf{bold}.}

\footnotesize
\setlength{\tabcolsep}{10pt}
\begin{tabular}{llccccc}
\toprule
\multicolumn{2}{c}{Methods} &Representation& Sedan & Bus/Truck & Pedestrian & mAP \\
\midrule
\multirow{3}{*}{\rotatebox[origin=c]{90}{Sup.}} 
 & \multirow{3}{*}{PTv3\cite{ptv3}}  
 &Point Cloud & 46.9/41.5 & 26.7/20.6 & 0.5/0.1 & 24.7/20.8\\
&&RAE& 51.2/42.9 & 28.4/23.8& 1.5/0.8& 27.0/22.5\\
&&RAD&49.2/41.5&23.8/17.4&18.7/14.9&30.6/24.6\\
\midrule
\multirow{7}{*}{\rotatebox[origin=c]{90}{Fine-tune}}  
  & +OccupancyMAE\cite{min2023occupancy} &Point Cloud &47.7/42.1 & 25.1/19.0 & 2.6/2.4 & 25.2/21.2\\  
  & +SONATA\cite{wu2025sonata} &Point Cloud & 49.2/42.5 & 29.2/23.2& 1.8/1.2&26.8/22.3 \\
  & +NOMAE \cite{nomae}&Point Cloud &  48.7/42.6 & 28.5/23.4 & 1.6/1.1 & 26.3/22.4  \\
  \cmidrule(lr){2-7}
    & \textbf{+SPMAE(Ours)}&RAE& 53.1/45.5 & 28.5/22.7 & 6.9/6.2 & 29.5/24.8  \\
  & \textbf{+L-SPMAE(Ours)}&RAE& 53.7/46.4 & 29.3/24.1 &5.0/4.2 & 29.3/24.9 \\
  & \textbf{+SPMAE(Ours)}&RAD& 53.5/45.8 & 31.3/\textbf{25.9} & 15.2/13.1 &33.4/28.2  \\
  & \textbf{+L-SPMAE(Ours)}&RAD& \textbf{54.0/47.3}& \textbf{33.3}/21.8 &\textbf{21.9/18.5} & \textbf{36.4/29.2} \\
\bottomrule
\end{tabular}

\label{tab: split2}
\end{table}
 Beyond the official split, we evaluate on a more challenging sequence split, as shown in Table~\ref{tab: split2}. Overall performance decreases due to stricter sequence-level separation, which requires stronger generalization. Both point clouds and 3D cubes struggle on the Pedestrian class, with or without SSL. In contrast, the 4D cube with Doppler-aware encoding achieves substantially better performance under both supervised and fine-tuning settings, highlighting the importance of effectively modeling Doppler information.

\subsection{Probing Comparison}
We provide probing comparisons on both the official K-Radar split (Table~\ref{tab: split1_probing}) and the sequence split (Table~\ref{tab: split2_probing}). On the official split, spectrum-based SSL consistently outperforms its point-cloud-based counterparts. In particular, L-SPMAE achieves gains of $(+3.4/+3.7)$ over SPMAE, demonstrating the effectiveness of the proposed leakage modeling during pretraining. On the sequence split, although NOMAE achieves competitive probing performance on Sedan and Bus/Truck categories, spectrum-based SSL shows clear superiority on Pedestrian and achieves the best overall mAP across all three classes. L-SPMAE also consistently outperforms SPMAE under this setting. Overall, these results demonstrate the advantage of spectrum-based SSL and validate the effectiveness of the proposed spectral leakage modeling strategy.

\begin{table}[h]
\centering
\caption{Probing comparison of \apbev/\apthd on the K-Radar official split.
The best performance is highlighted in \textbf{bold}.}

\footnotesize
\setlength{\tabcolsep}{10pt}
\begin{tabular}{llccccc}
\toprule
\multicolumn{2}{c}{PTv3} &Representation& Sedan & Bus/Truck & Pedestrian & mAP\\
\midrule
\multirow{5}{*}{\rotatebox[origin=c]{90}{Probing}}  
 & +OccupancyMAE\cite{min2023occupancy} &Point Cloud &  55.2/49.0&43.1/34.9 &38.8/35.8 &45.7/39.8\\
 & +SONATA\cite{wu2025sonata} &Point Cloud & 55.8/49.7 &41.3/31.9  &38.0/36.5  &45.0/39.4  \\  
  & +NOMAE\cite{nomae}&Point Cloud &56.0/49.9  &43.6/33.4 & 38.2/35.8&46.0/39.7 \\
  \cmidrule(lr){2-7}
  & \textbf{+SPMAE(Ours)}&RAD &56.6/49.9&46.2/36.7&41.5/38.5&48.1/41.7\\
  & \textbf{+L-SPMAE(Ours)}&RAD & \textbf{59.2/52.6}&\textbf{51.2/41.7}&\textbf{44.1}/\textbf{41.9}&\textbf{51.5/45.4} \\
\bottomrule
\end{tabular}

\label{tab: split1_probing}
\end{table}
\begin{table}[t]
\centering
\caption{Probing comparison of \apbev/\apthd on the K-Radar sequence split.
The best performance is highlighted in \textbf{bold}.}
\footnotesize
\setlength{\tabcolsep}{10pt}
\begin{tabular}{llccccc}
\toprule
\multicolumn{2}{c}{PTv3} &Representation& Sedan & Bus/Truck & Pedestrian & mAP \\
\midrule
\multirow{5}{*}{\rotatebox[origin=c]{90}{Probing}}  
  & +OccupancyMAE\cite{min2023occupancy} &Point Cloud & 45.5/40.1 & 17.8/13.6 & 4.4/2.8 & 22.6/18.9\\  
  & +SONATA\cite{wu2025sonata} &Point Cloud & 45.2/39.8 & 14.3/10.2& 3.8/3.5 & 21.1/17.8 \\  
  & +NOMAE\cite{nomae} &Point Cloud & 48.0/\textbf{42.5}& \textbf{21.7/12.3}& 4.3/3.9& 24.7/19.6 \\
  \cmidrule(lr){2-7}
  & \textbf{+SPMAE(Ours)}&RAD& 46.4/39.6 &15.9/10.8  &20.3/17.4  &27.5/22.6  \\
  & \textbf{+L-SPMAE(Ours)}&RAD& \textbf{49.3}/42.3 & 16.9/11.4 &\textbf{22.3/19.0}&\textbf{29.5/24.2} \\
\bottomrule
\end{tabular}
\label{tab: split2_probing}
\end{table}

\subsection{Multi-object Tracking}
\label{sec: Tracking}
We further evaluate the transferability of the learned representations on radar-based multi-object tracking, following the official K-Radar tracking protocol\footnote{\url{https://github.com/kaist-avelab/K-Radar/blob/main/docs/tracking.md}}. As shown in Table~\ref{tab:tracking_results}, both SPMAE and L-SPMAE outperform training from scratch, demonstrating effective transfer beyond object detection. L-SPMAE achieves the best AMOTA, MOTA, and recall, while SPMAE obtains a slightly better AMOTP.
\begin{table}[t]
\centering
\caption{Radar-based multi-object tracking results on K-Radar. We use SimpleTrack~\citep{pang2022simpletrack} as the tracking framework.}

\begin{tabular}{lcccc}
\toprule
Methods & AMOTA $\uparrow$ & MOTA $\uparrow$ & Recall $\uparrow$ & AMOTP $\downarrow$ \\
\midrule
PTv3& 0.529 & 0.486 & 0.567 & 1.038 \\
\midrule
\textbf{+SPMAE(Ours)} & 0.562 & 0.517 & 0.607 & \textbf{0.973} \\
\textbf{+L-SPMAE(Ours)} & \textbf{0.577} & \textbf{0.530} & \textbf{0.640} & 0.975 \\
\bottomrule
\end{tabular}
\label{tab:tracking_results}
\small
\end{table}

\subsection{Label-Efficient Fine-tuning}
Table~\ref{tab:efficient} reports L-SPMAE fine-tuned with varying fractions of
labeled data. Pretraining consistently improves performance across all data scales,
and fine-tuning with only $25\%$ of labels already surpasses the fully supervised
baseline trained on $100\%$ of the data, demonstrating that signal discovery
substantially reduces annotation requirements, which is particularly valuable for
radar, where labeling is costly. Our pretraining is complementary to semi-supervised
approaches that exploit unlabeled data during task training~\cite{yang2026collis},
and combining the two is a promising direction for future work.

\begin{table}[tb]
    \centering
    \captionof{table}{Fine-tuning mAP on the K-Radar official split with different annotated subsets.}
    \label{tab:efficient}
    \footnotesize
\begin{tabular}{lccccc}
\toprule
Ratio ($\%$)& 10 & 25 & 50&75&100 \\
\midrule
From Scratch & 12.7/8.9&42.2/35.7&43.5/37.9&45.4/38.9&48.9/43.1 \\
+L-SPMAE&34.5/28.1&50.4/44.4&52.4/47.0&54.5/48.5& 55.9/50.1\\
\bottomrule
\end{tabular}

\end{table}
\subsection{Qualitative Results}
\begin{figure*}[tb]
    \centering
    \includegraphics[width=0.95\linewidth]{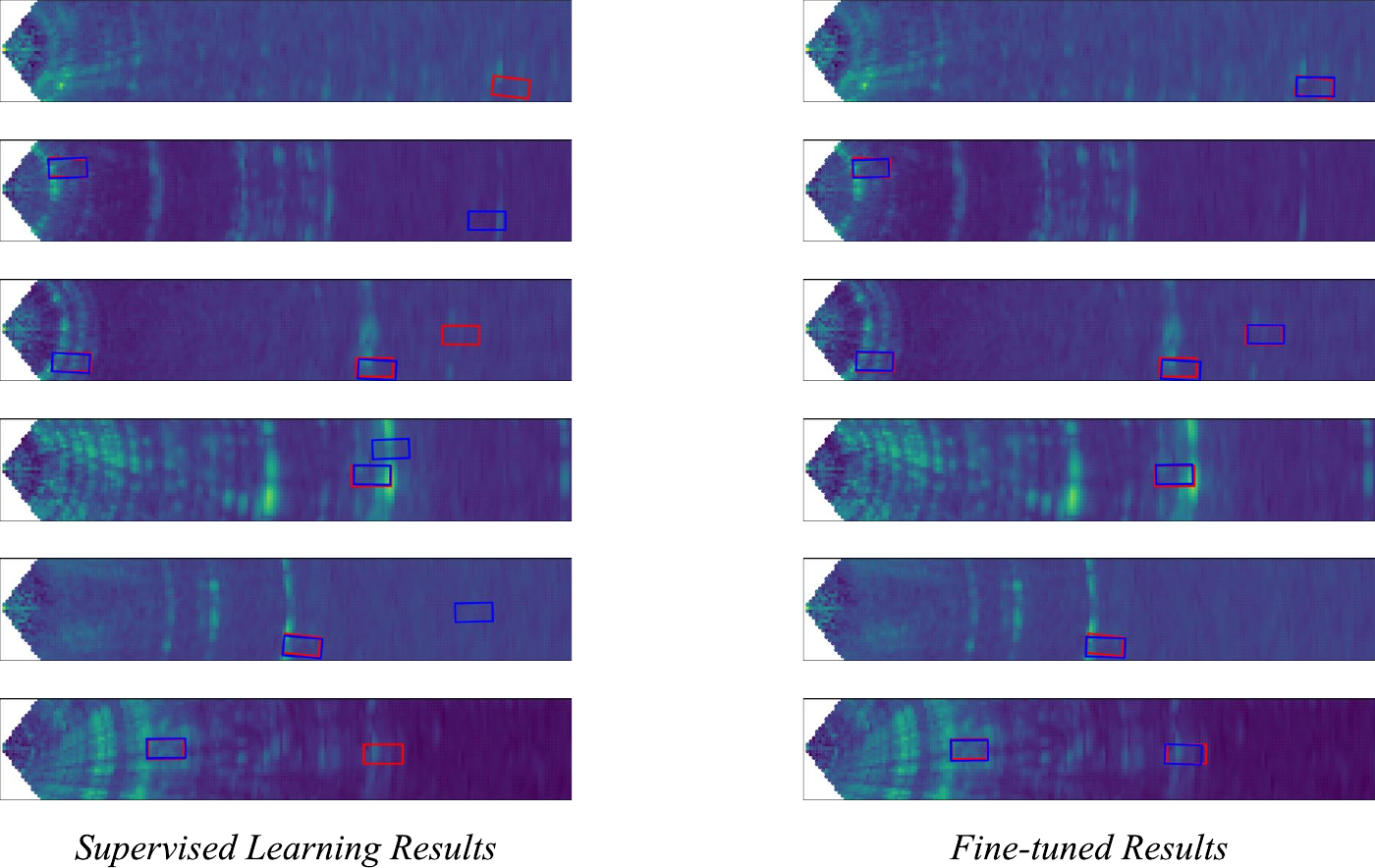}
        \caption{Prediction results on Bird’s-eye view.
(\textcolor{red}{ground truth}; \textcolor{blue}{prediction}).}
\label{fig:vis}
\end{figure*}
Figure~\ref{fig:vis} gives qualitative results of how our L-SPMAE pretraining benefits the downstream task. Generally, the pretraining helps the networks better exploit spectral features, providing a more accurate prediction, particularly at long range.


\section{Running Environment}
\label{appendix: running_environment}

All experiments are conducted on a single NVIDIA B200 GPU. Our implementation is based on Pointcept~\cite{pointcept2023}, MMDetection3D~\cite{mmdet3d2020}, and OpenPCDet~\cite{openpcdet2020}, using PyTorch 2.8, spconv 2.x, and FlashAttention2. Since FlashAttention2 requires GPUs with SM80 or newer architectures, Ampere or later generations are needed for reproduction. With a batch size of 8, our method requires at least 50\,GB GPU memory for training and 13\,GB for inference. The code will be publicly released after internal approval.
\section{Running Efficiency}
\label{app: efficiency}
Although the radar spectrum preserves substantially richer information than sparse point clouds and can benefit perception performance, it also introduces significantly higher computational cost. In particular, the raw RAE cube is approximately \(10\times\) larger in data volume than the corresponding sparse point cloud representation. Table~\ref{tab: speed} compares the running speed on the same GPU device. Compared with point-cloud-based processing, 3D spectrum processing is \(3.9\times\) slower during training and \(4.8\times\) slower during inference, while 4D spectrum processing is \(9.5\times\) slower during training and \(11\times\) slower during inference. In future work, we plan to explore token pruning techniques~\cite{gitmerge3d2025} for spectrum representations to further reduce the computational cost.
\begin{table}[t]
\centering
\caption{Comparison of running speed with PTv3 for object detection on a single NVIDIA B200 GPU, where each iteration uses a batch size of 8. Note that the reported speed is not the upper performance bound, since the implementation is not fully hardware-optimized. }
\begin{tabular}{lcc}
\toprule
Representation & Training iterations/s &Inference iterations/s \\
\midrule
Point Cloud & 6.7&14.3 \\
3D Spectrum  & 1.7&3.0 \\
4D Spectrum & 0.7&1.3 \\

\bottomrule
\end{tabular}

\label{tab: speed}
\end{table}
\section{Algorithm Details}
\label{appendix: algo}
The details of spectral leakage modeling are illustrated in Algorithm~\ref{alg: spectral_leakage}. We first map the spectrum points into a generalized 3D spectral grid for modeling the leakage effect directly in the spectral domain. The three spectral axes are not restricted to a specific representation and can correspond to different coordinate parameterizations, such as RAE or RAD. To formulate the 3D leakage kernel, we construct three normalized 1D leakage kernels from the block-wise masks along each spectral axis. In practice, the convolution is efficiently implemented as $\mathrm{IFFT}\big(\mathrm{FFT}(S)\odot \mathrm{FFT}(K)\big)$ according to the convolution theorem. For training stability, we further rescale the corrupted spectrum to preserve the total spectral energy of the original spectrum.

\begin{algorithm}[H]
\caption{Spectral Leakage Modeling}
\label{alg: spectral_leakage}
\footnotesize
\begin{algorithmic}[1]
\Require Points $\{(\mathbf{x}_i,f_i)\}_{i=1}^P$; 
grid mapping $\pi(\cdot)$; grid size $(D_1,D_2,D_3)$; 
leakage rate $\lambda$; block size $b$
\Ensure Corrupted point powers $\{f'_i\}_{i=1}^P$

\Statex \textbf{Step 1: Map points to a generalized 3D spectral grid.}

\State Compute spectral-grid indices:
\Statex \[
\mathbf{u}_i \leftarrow \pi(\mathbf{x}_i),
\qquad
\mathbf{u}_i=(u_i^{(1)},u_i^{(2)},u_i^{(3)}).
\]

\State Scatter point powers to a dense spectral grid:
\Statex \[
S \in \mathbb{R}^{D_1\times D_2\times D_3},
\qquad
S[\mathbf{u}_i] \leftarrow S[\mathbf{u}_i] + f_i.
\]

\Statex \textbf{Step 2: Generate block-wise leakage masks.}

\For{$j=1,2,3$}
    \State Let $D_j$ denote the length of the $j$-th spectral axis.
    
    \State Sample a 1D block-wise binary mask:
    \Statex \[
    m_j \leftarrow \mathrm{BlockMask}(D_j,\lambda,b),
    \]
    
    \Statex where $\lambda$ controls the leakage strength and $b$ controls the spatial extent of corrupted regions.
\EndFor

\Statex \textbf{Step 3: Convert block masks to leakage kernels.}

\For{$j=1,2,3$}
    \State Compute the spectral power response:
    \Statex \[
    \bar{K}_j \leftarrow |\mathrm{FFT}(m_j)|^2.
    \]
    
    \State Normalize into a 1D leakage kernel:
    \Statex \[
    K_j \leftarrow \frac{\bar{K}_j}{\sum_{q=1}^{D_j}\bar{K}_j[q]}.
    \]
\EndFor

\Statex \textbf{Step 4: Construct a separable 3D leakage kernel.}

\State Form the separable kernel:
\Statex \[
K[\mathbf{u}]
\leftarrow
K_1[u^{(1)}]\cdot
K_2[u^{(2)}]\cdot
K_3[u^{(3)}].
\]

\Statex \textbf{Step 5: Apply leakage by FFT-based convolution.}

\State Apply circular convolution:
\Statex \[
\tilde{S}
\leftarrow
\Re\left\{
\mathrm{IFFT}
\left(
\mathrm{FFT}(S)\odot \mathrm{FFT}(K)
\right)
\right\}.
\]

\Statex \textbf{Step 6: Preserve total spectral energy.}

\State Compute spectral energies:
\Statex \[
E_{\mathrm{in}} \leftarrow \sum_{\mathbf{u}} S[\mathbf{u}],
\qquad
E_{\mathrm{out}} \leftarrow \sum_{\mathbf{u}} \tilde{S}[\mathbf{u}].
\]

\State Calibrate the leaked spectrum:
\Statex \[
\tilde{S}
\leftarrow
\frac{E_{\mathrm{in}}}{E_{\mathrm{out}}+\epsilon}\tilde{S}.
\]

\Statex \textbf{Step 7: Gather corrupted values back to points.}

\State Assign corrupted powers:
\Statex \[
f'_i \leftarrow \tilde{S}[\mathbf{u}_i].
\]

\State \Return $\{f'_i\}_{i=1}^P$

\end{algorithmic}
\end{algorithm}
\section{Broader Impact}
\label{app: broader impact}
This work investigates self-supervised learning on dense radar spectrum representations for 3D perception. The proposed method may improve the robustness of radar-based perception systems for safety-critical applications such as autonomous driving, especially under adverse weather conditions. Potential risks include incorrect detections in downstream autonomous systems and possible misuse in surveillance scenarios. These risks can be mitigated through careful validation and evaluation before deployment.




\newpage

\end{document}